\documentclass[runningheads]{llncs}
\usepackage[T1]{fontenc}
\usepackage{graphicx}
\usepackage{booktabs}
\usepackage{multirow}
\usepackage{bm}
\usepackage{cite}
\usepackage{url}
\usepackage{orcidlink}
\begin{document}
\title{Interpretability-Aware Pruning for Efficient Medical Image Analysis}
\author{
Nikita Malik\inst{1,2}\orcidlink{0009-0006-1253-6269}\and
Pratinav Seth\inst{1}\orcidlink{0009-0001-4525-4464}\and
Neeraj K. Singh\inst{1}\orcidlink{https://orcid.org/0009-0006-5465-4247}\and \\
Chintan Chitroda\inst{1}\and
Vinay K. Sankarapu\inst{1}\orcidlink{https://orcid.org/0000-0002-9416-3497}
}
\authorrunning{N. Malik et al.}
\institute{AryaXAI Alignment Lab, AryaXAI.com, Mumbai, India \and
Manipal Institute of Technology, Manipal, India\\
\email{pratinav.seth@aryaxai.com}}
\maketitle
\begin{abstract}
Deep learning has driven significant advances in medical image analysis, yet its adoption in clinical practice remains constrained by the large size and lack of transparency in modern models. Advances in interpretability techniques such as DL-Backtrace, Layer-wise Relevance Propagation, and Integrated Gradients make it possible to assess the contribution of individual components within neural networks trained on medical imaging tasks. In this work, we introduce an interpretability-guided pruning framework that reduces model complexity while preserving both predictive performance and transparency. By selectively retaining only the most relevant parts of each layer, our method enables targeted compression that maintains clinically meaningful representations. Experiments across multiple medical image classification benchmarks demonstrate that this approach achieves high compression rates with minimal loss in accuracy, paving the way for lightweight, interpretable models suited for real-world deployment in healthcare settings\footnote{Code will be available at : 

https://github.com/AryaXAI/interpretability-aware-pruning-healthcare}.
\end{abstract}
\section{Introduction}
Deep learning has revolutionized numerous fields, with medical image analysis being a prominent beneficiary \cite{litjens2017survey, shen2017deep}. Convolutional Neural Networks (CNNs) and other deep architectures have demonstrated state-of-the-art performance across a spectrum of tasks, from disease detection and segmentation to diagnosis and prognosis. Despite their remarkable accuracy, the widespread adoption of these models in clinical settings is often hampered by two critical limitations: their inherent computational complexity and their "black-box" nature \cite{esteva2021deep, holzinger2017we}.
The former necessitates significant computational resources, making deployment challenging in environments with limited infrastructure, while the latter hinders trust and understanding among medical professionals who require transparent and explainable insights into model decisions.

To address these challenges, model compression techniques, particularly pruning, have emerged as a promising avenue \cite{blalock2020state, liu2019rethinking}. Pruning aims to reduce the size and computational footprint of neural networks by removing redundant or less important parameters.
Traditionally, pruning methods have relied on heuristics such as weight magnitude or Random removal \cite{han2015learning, frankle2019lottery, gale2019state}.
While effective in reducing model size, these approaches often lack a direct connection to the model's decision-making process, potentially discarding parameters that contribute to critical features or interpretability.

This paper introduces an interpretability-aware pruning method that leverages advanced attribution techniques to guide the compression process. By integrating methods like Layer-wise Relevance Propagation (LRP) \cite{montavon2017explaining}, DL Backtrace (DLB) \cite{sankarapu2025dlbacktracemodelagnosticexplainability}, and Integrated Gradients (IG) \cite{sundararajan2017axiomatic}, we compute importance scores for individual layers (and neurons), reflecting their contributions to the model's prediction. These scores then inform a targeted pruning strategy, allowing us to remove less relevant components while preserving the model's performance and enhancing its transparency. Our approach is particularly effective in identifying and removing individual neurons that are not crucial for model prediction, leading to significant lossless compression. We demonstrate the efficacy of our method on diverse medical imaging datasets, showcasing its potential to create efficient and interpretable deep learning models for real-world clinical applications.

\section{Related Works}

\subsection{Model Pruning}
Model pruning is a widely adopted technique for neural network compression, aiming to reduce model size and accelerate inference by eliminating redundant parameters.
Pruning methods can be broadly categorized based on when and how the pruning is applied. Train-time pruning integrates the pruning process directly into the training phase \cite{zhu2018prune, sanh2020movement}, while post-training pruning, conversely, applies pruning techniques to a fully trained model as a separate step \cite{blalock2020state, hassibi1993second}.
Our work focuses on a post-training approach where importance scores are computed on a trained model for individual layers and neurons \cite{hatefi2024pruningexplainingrevisitedoptimizing}.
It focuses on identifying and pruning individual neurons, which aligns with unstructured pruning at a conceptual level, but with an interpretability-guided selection.

\subsection{Model Interpretability and Model Compression}

Model Interpretability aims to make the decision-making processes of complex machine learning models more transparent and understandable. In critical domains such as healthcare, finance, and law enforcement, where AI-driven decisions can have significant ethical, legal, and societal implications, the ability to explain how AI systems arrive at their conclusions is paramount~\cite{arrieta2019explainableartificialintelligencexai, gohel2021explainableaicurrentstatus, atakishiyev2025safetyimplicationsexplainableartificial, fresz2024contributionxaisafedevelopment}.
Beyond enhancing transparency and trust, Model Interpretability can also facilitate model compression. By identifying which components of a model—such as specific neurons, filters, or layers—are most influential in its predictions, it is possible to remove less critical parts without substantially affecting performance. This approach, often termed "pruning by explaining," leverages interpretability to guide the simplification of models, making them more efficient and suitable for deployment in resource-constrained environments~\cite{sabih2020utilizing}.

To implement interpretability-aware pruning, we utilize several techniques that assess the contribution of different components within a neural network:

\textbf{Layer-wise Relevance Propagation (LRP):} LRP operates by propagating the prediction backward through the network, distributing a "relevance score" from the output layer down to the input features or intermediate neurons. A key property of LRP is its conservation principle, ensuring that the total relevance is preserved across layers. This allows for a quantitative measure of how much each neuron contributes to the final prediction, making it suitable for identifying less important components~\cite{Montavon2019LayerWiseRP}.

\textbf{DL-Backtrace (DLB):} DL-Backtrace is a model-agnostic method that assigns relevance scores across layers, revealing feature importance, information flow, and potential biases in predictions. A significant advantage of DLB is its independence from auxiliary models or baselines, ensuring consistent and deterministic interpretations across diverse architectures (MLPs, CNNs, LLMs) and data types (images, text, tabular data). This makes it a robust tool for our pruning strategy~\cite{sankarapu2025dlbacktracemodelagnosticexplainability}.

\textbf{Integrated Gradients (IG):} IG is a path-based explanation technique that quantifies the contribution of individual input features to a model's prediction. It works by integrating the gradients of the model output with respect to the input features along a straight-line path from a baseline input (e.g., a black image) to the actual input. IG is known for satisfying desirable axioms and can mitigate issues like saturation effects, providing less noisy and more complete explanations~\cite{Sundararajan2017AxiomaticAF}.

By applying these interpretability techniques, we can derive importance scores that inform our pruning strategy. This allows us to move beyond simple magnitude-based heuristics and instead focus on the functional importance of model components, leading to more efficient and interpretable models.

\section{Methodology}
\begin{figure}[pt]
    \centering
    \includegraphics[width=0.81\linewidth]{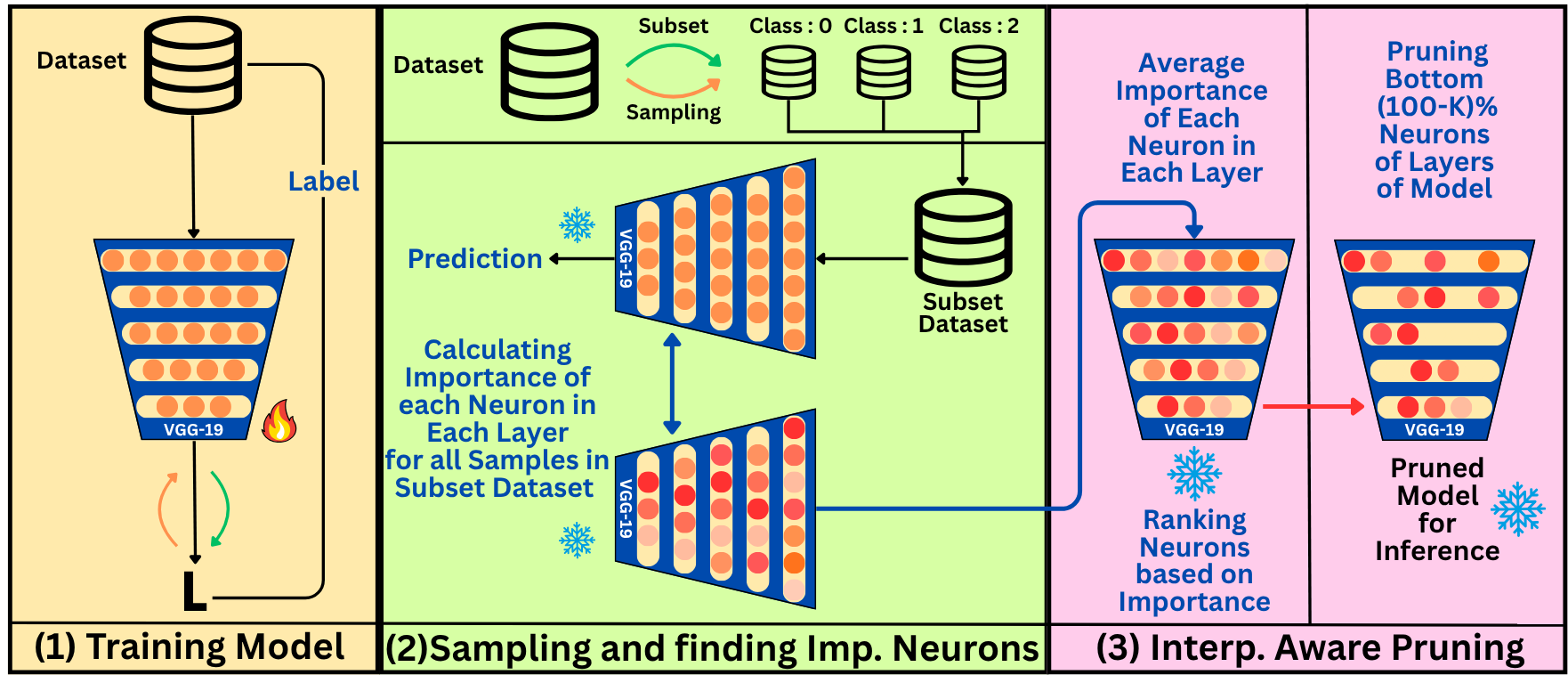}
    \caption{Workflow for Interpretability-Aware Model Pruning: (1) Train the model on the dataset with labels, (2) Perform sampling and rank neurons based on their importance in each layer using subset datasets, (3) Prune the least important neurons (bottom (100-K)\%) from each layer to create a compressed model for inference.}
    \label{fig:fig1}
\end{figure}

Our proposed interpretability-aware pruning method integrates interpretability techniques with a targeted neuron removal strategy to achieve efficient and transparent model compression. The entire pruning process (as illustrated in Figure~\ref{fig:fig1}) involves several key steps, from computing importance scores to iteratively pruning and evaluating the model.

\subsection{Importance Score Computation}
The foundation of our method lies in accurately quantifying the importance of individual neurons within the deep learning model. For this, we employ three distinct attribution methods: Layer-wise Relevance Propagation (LRP), DL Backtrace (DLB), and Integrated Gradients (IG). Each of these methods produces scores at the initial and intermediate feature levels. To approximate importance at the neuron level, we aggregate these importance values by summing across the channel dimension to obtain a per-neuron importance score aligned with the output of each feature map.
This aggregation step helps to smooth out noise and provide a more stable measure of a neuron's overall importance.

\subsection{Sample Selection for Importance Computation}
To ensure that the computed importance scores are representative and robust, we utilize various sampling techniques to select 10 samples per class from the training set on which the importance scores are calculated. This addresses the potential for importance scores to vary across different inputs. The sampling techniques explored include:

\textbf{Confidence Sampling}: Selects samples for which the model has the highest prediction Confidence. This focuses on inputs where the model's decision is strong, potentially highlighting the most critical neurons.

\textbf{Random Sampling}: A Random selection of samples from each class is used. This provides a general overview of neuron importance across the dataset.

\textbf{Clustering-based Sampling}: Samples are grouped into clusters based on their features or model activations, and representatives from each cluster are selected. This ensures diversity in the chosen samples, covering different data distributions.

\subsection{Pruning Strategy}
Once the aggregated scores for all neurons are obtained, we proceed with the pruning process by identifying and removing the "bottom-most" neurons, i.e., those with the lowest scores, up to a specified pruning threshold. For each layer or the entire network, neurons are ranked across the entire network based on their aggregated scores. A predefined pruning rate determines the percentage of neurons with the lowest importance scores to be removed. Neurons falling below this threshold are considered low-importance and are pruned by setting their associated weights to zero. This effectively eliminates their contribution to the network's computation, aligning with an unstructured pruning strategy at the neuron level, driven by interpretability.

\section{Experimental Setup}
\subsection{Datasets and Models}
We utilized four medical imaging datasets, each with unique clinical contexts:

\textbf{MURA Dataset} \cite{rajpurkar2017mura}: A large-scale dataset of musculoskeletal radiographs from Stanford, comprising seven standard study types (e.g., elbow, wrist, shoulder), with each X-ray labeled as normal or abnormal.

\textbf{KVASIR Dataset} \cite{pogorelov2017kvasir} : KVASIR is a collection of  Endoscopic images of the gastrointestinal tract, categorized by anatomical landmarks (e.g., Z-line, pylorus) and pathological findings (e.g., esophagitis, polyps).

\textbf{CPN Dataset \cite{Kumar2022-gg}:} The CPN (Common Peroneal Nerve) dataset focuses on diagnosing common peroneal nerve injuries in the lower limb. This dataset is used for classification tasks in nerve injury detection and related clinical scenarios.

\textbf{FETAL Planes Dataset \cite{Burgos-Artizzu2020-ly}:} This dataset comprises high-resolution ultrasound images of fetal heads, highlighting regions such as the brain, cavum septum pellucidum (CSP), and lateral ventricles (LV), supporting both classification and detection tasks.

We applied our pruning methodology across three widely used architectures: VGG19 \cite{simonyan2014deep}, ResNet50 \cite{he2016deep}, and Vision Transformer (ViT) \cite{dosovitskiy2020image}. Each model was trained in a supervised fashion on all datasets to establish a strong baseline. This process incorporated data augmentation techniques, learning rate schedulers, optimization algorithms, and loss functions, including Cross Entropy, to achieve optimal results.

\subsection{Pruning Process and Evaluation}
For each model and dataset, we followed the methodology outlined in Section 3. Importance scores were aggregated from a subset of samples obtained from the specified sampling techniques (most Confidence, Random, Clustering with 10 samples per class from the training set). Neurons with the lowest aggregated importance scores were pruned based on a defined pruning rate. We evaluated the models' accuracy and computational efficiency at various pruning rates, specifically observing the point of "lossless pruning" where accuracy is maintained despite significant compression.
Although the importance score computation incurs a one-time overhead proportional to model size, this cost is theoretically justified by the resulting gains in post-pruning inference efficiency, thereby amplifying benefits in resource-limited clinical settings.

\section{Results and Discussion}
\begin{figure}[!h]
    \centering
    \includegraphics[width=0.9\textwidth,height=\textheight,keepaspectratio]{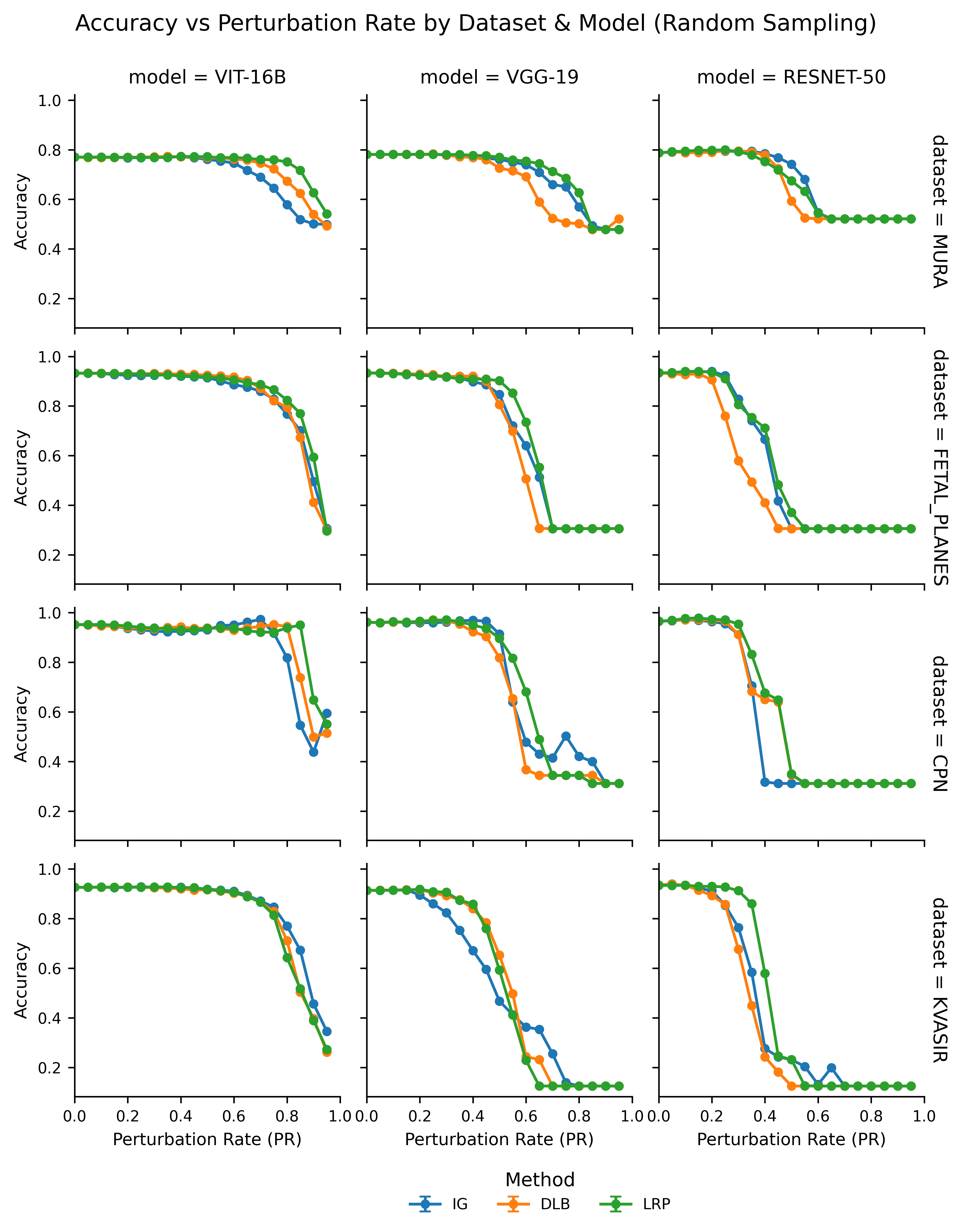}
    \caption{Accuracy vs Pruning Rate across various Models (Vit,Resnet,Vgg) and Datasets (Mura,Fetal Planes,Cpn,Kvasir) using Various Attribution Methods (IG,DLB,LRP).}
    \label{fig:pruning_comparison}
\end{figure}
\begin{table}[pt]
\centering
\caption{Comparison of accuracy drop from unpruned model at different pruning rates averaged across all datasets and model architectures for various Interpretability Methods (DLB, IG , LRP) and Sampling Methods (Clustering, Confidence, Random).}
\label{tab:my-table}
\footnotesize
\begin{tabular}{cccccc}
\hline
\multirow{2}{*}{Method} & \multirow{2}{*}{Sampling} & \multicolumn{4}{c}{$\Delta (\downarrow) \mathrm{Acc} $ Drop from Baseline at Pruning Rate}     \\
                        &                           & 0.15                   & 0.30                   & 0.50                   & 0.70                   \\ \hline
DLB                     & Clustering                & $\bm{0.001 \pm 0.01}$  & $0.058 \pm 0.11$      & $0.234 \pm 0.30$      & $\bm{0.384 \pm 0.31}$ \\
DLB                     & Confidence                & $0.002 \pm 0.01$       & $\bm{0.054 \pm 0.11}$ & $\bm{0.215 \pm 0.27}$ & $0.398 \pm 0.31$      \\
DLB                     & Random                    & $0.002 \pm 0.01$       & $0.059 \pm 0.12$      & $0.239 \pm 0.28$      & $0.400 \pm 0.32$      \\ \hline
IG                      & Clustering                & $\bm{0.000 \pm 0.00}$  & $0.036 \pm 0.07$      & $0.255 \pm 0.29$      & $0.389 \pm 0.31$      \\
IG                      & Confidence                & $0.002 \pm 0.01$       & $\bm{0.026 \pm 0.04}$ & $0.253 \pm 0.27$      & $0.398 \pm 0.30$      \\
IG                      & Random                    & $0.002 \pm 0.01$       & $0.039 \pm 0.06$      & $\bm{0.224 \pm 0.29}$ & $\bm{0.375 \pm 0.30}$ \\ \hline
LRP                     & Clustering                & $\bm{-0.001 \pm 0.00}$ & $0.018 \pm 0.04$      & $\bm{0.196 \pm 0.28}$ & $\bm{0.382 \pm 0.33}$ \\
LRP                     & Confidence                & $0.000 \pm 0.00$       & $0.027 \pm 0.04$      & $0.203 \pm 0.28$      & $0.397 \pm 0.31$      \\
LRP                     & Random                    & $\bm{-0.001 \pm 0.00}$ & $\bm{0.016 \pm 0.04}$ & $0.205 \pm 0.27$      & $0.384 \pm 0.33$      \\ \hline
\end{tabular}
\end{table}
The results demonstrate the effectiveness of the proposed pruning methodology across different medical imaging datasets and model architectures. This section highlights core findings related to model resilience under high pruning rates, the impact of attribution methods, and the influence of sampling strategies.
Table~\ref{tab:my-table} and Figure~\ref{fig:pruning_comparison} summarize pruning performance across the experimental spectrum. Across all datasets and models, interpretability-guided pruning using LRP, DLB, and IG consistently enables high model compression while maintaining accuracy within a 5\% margin, even at moderate pruning rates.
\subsection{Model-Wise Observations}
We observed across all datasets that, in most cases, ViTs seem to have the highest accuracy followed by Vgg and Resnet.
Among the architectures, ViTs demonstrate exceptional robustness, consistently preserving performance even beyond 65–80\% pruning. Notably, in certain configurations (e.g., CPN with LRP or DLB), ViTs withstand pruning rates of 85\% with minimal degradation. This performance suggests a higher degree of redundancy or feature richness in transformer-based models compared to CNNs.
VGG architectures typically tolerate up to 50\% pruning with less than 5\% accuracy drop, whereas ResNets are comparatively sensitive, maintaining stability only up to 30\% pruning in most settings. The observed variation highlights the architectural differences: residual connections in ResNet may cause greater entanglement of information across neurons, making unstructured neuron pruning comparatively less impactful.

\subsection{Impact of different Interpretability Methods}
LRP consistently achieves the highest pruning thresholds while maintaining accuracy. As shown in the pruning rate vs. accuracy plots, LRP also achieves the lowest accuracy drop across pruning rates, slightly better than IG and DLB. This affirms LRP’s strong conservation properties and robustness in identifying less important neurons followed by DLB and IG.

\subsection{Impact of different sampling strategies}
Clustering-based sampling performs best on average across attribution methods, particularly noticeable at moderate pruning levels (30–50\%). As shown in Table \ref{tab:my-table}, LRP with Clustering shows the lowest drop, outperforming Random and Confidence-based alternatives. Confidence-based sampling performs competitively, particularly for IG and DLB, where it occasionally surpasses Random selection in maintaining model stability.
Interestingly, in certain configurations (especially at lower pruning rates like 15–30\%), a slight increase in accuracy is observed. This counterintuitive behavior arises because pruning also removes neurons that may have a negative or noisy influence on predictions. By discarding these misleading neurons, the model generalizes better, resulting in a net accuracy gain.
The sampling strategy also influences pruning stability; while minor fine-tuning may be required for specific datasets, the overall framework remains robust across sampling choices.
The observed high variance is expected as these results are averaged across different datasets and architectures; however, overall trends remain consistent.

\section{Conclusion}
In this work, we introduce an interpretability-aware pruning framework that leverages multiple interpretability techniques to guide the compression of deep neural networks used in medical image analysis. Instead of relying on traditional magnitude-based or heuristic pruning strategies, our method utilizes importance scores derived from model interpretability techniques, enabling a principled and transparent approach to model compression.

We evaluate our framework on four diverse medical imaging datasets—MURA, CPN, KVASIR, and Fetal Planes—across three representative architectures: VGG19, ResNet50, and VIT-16b. Our experiments demonstrate that the proposed method achieves substantial pruning rates (up to 80–85\%) with minimal loss in accuracy. Notably, ViT-based models retained performance particularly well under compression, highlighting their potential for deployment in resource-constrained clinical environments.
Among the interpretability techniques studied, Layer-wise Relevance Propagation (LRP) consistently enabled the highest pruning thresholds without degrading performance, followed by DL-Backtrace (DLB) and Integrated Gradients (IG). We also observed that the choice of sampling strategy significantly impacts pruning effectiveness: clustering-based sampling outperformed Confidence-based and Random strategies, achieving the most robust results.
Interestingly, in certain settings, we observed improved accuracy after pruning, suggesting that interpretability-guided pruning may help eliminate redundant or noisy neurons, thereby improving generalization. This reinforces its dual role in facilitating both compression and performance enhancement.

Overall, our findings highlight that interpretability-guided pruning offers an effective pathway to building lightweight, interpretable, and potentially more robust AI systems—an essential step toward the safe and trustworthy deployment of AI systems in high-stakes domains such as healthcare.

\section{Implications for Medical Image Analysis}
Interpretability-aware pruning offers significant benefits for healthcare:
\begin{itemize}
    \item \textbf{Efficiency:} Achieving 40--50\% lossless pruning creates lightweight models with reduced memory and computation needs, enabling deployment on edge devices, mobile platforms, and clinics with limited resources, while speeding up diagnostic workflows.
    \item \textbf{Interpretability:} Pruning guided by interpretability methods links model compression with transparency, highlighting which neurons are critical and helping clinicians understand model decisions based on meaningful features.
    \item \textbf{Robustness:} Using diverse sampling methods to evaluate neuron importance ensures pruning preserves neurons important for varied inputs and edge cases, enhancing model reliability.
\end{itemize}
Although further empirical validation is needed, we hypothesize that pruning pre-trained models using region-specific, representative data could improve generalization at deployment. This approach may mitigate domain shift and preserve model reliability in real-world clinical settings.

\section*{Disclosure of Interests}
The authors are affiliated with AryaXAI.com, part of Aurionpro Solutions Limited. DL-Backtrace was introduced by AryaXAI. The authors declare no further competing interests.

\bibliography{ref}

\begin{thebibliography}{10}
\providecommand{\url}[1]{\texttt{#1}}
\providecommand{\urlprefix}{URL }
\providecommand{\doi}[1]{https://doi.org/#1}

\bibitem{arrieta2019explainableartificialintelligencexai}
Arrieta, A.B., Díaz-Rodríguez, N., Ser, J.D., Bennetot, A., Tabik, S., Barbado, A., García, S., Gil-López, S., Molina, D., Benjamins, R., Chatila, R., Herrera, F.: Explainable artificial intelligence (xai): Concepts, taxonomies, opportunities and challenges toward responsible ai (2019), \url{https://arxiv.org/abs/1910.10045}

\bibitem{atakishiyev2025safetyimplicationsexplainableartificial}
Atakishiyev, S., Salameh, M., Goebel, R.: Safety implications of explainable artificial intelligence in end-to-end autonomous driving (2025), \url{https://arxiv.org/abs/2403.12176}

\bibitem{blalock2020state}
Blalock, D., Ortiz, J.J.G., Frankle, J., Guttag, J.V.: What is the state of neural network pruning? Proceedings of Machine Learning and Systems  \textbf{2},  129--146 (2020)

\bibitem{Burgos-Artizzu2020-ly}
Burgos-Artizzu, X.P., Coronado-Guti{\'e}rrez, D., Valenzuela-Alcaraz, B., Bonet-Carne, E., Eixarch, E., Crispi, F., Gratac{\'o}s, E.: Evaluation of deep convolutional neural networks for automatic classification of common maternal fetal ultrasound planes. Sci. Rep.  \textbf{10}(1),  10200 (Jun 2020)

\bibitem{dosovitskiy2020image}
Dosovitskiy, A., Beyer, L., Kolesnikov, A., Weissenborn, D., Zhai, X., Unterthiner, T., Dehghani, M., Minderer, M., Heigold, G., Gelly, S., Uszkoreit, J., Houlsby, N.: An image is worth 16x16 words: Transformers for image recognition at scale. In: International Conference on Learning Representations (ICLR) (2021)

\bibitem{esteva2021deep}
Esteva, A., Chou, K., Yeung, S., Naik, N., Madani, A., Mottaghi, A., Liu, Y., Topol, E., Dean, J., Socher, R.: Deep learning-enabled medical computer vision. npj Digital Medicine  \textbf{4}(1), ~5 (2021)

\bibitem{frankle2019lottery}
Frankle, J., Carbin, M.: The lottery ticket hypothesis: Finding sparse, trainable neural networks. In: International Conference on Learning Representations (ICLR) (2019)

\bibitem{fresz2024contributionxaisafedevelopment}
Fresz, B., Göbels, V.P., Omri, S., Brajovic, D., Aichele, A., Kutz, J., Neuhüttler, J., Huber, M.F.: The contribution of xai for the safe development and certification of ai: An expert-based analysis (2024), \url{https://arxiv.org/abs/2408.02379}

\bibitem{gale2019state}
Gale, T., Elsen, E., Hooker, S.: The state of sparsity in deep neural networks. arXiv preprint arXiv:1902.09574  (2019)

\bibitem{gohel2021explainableaicurrentstatus}
Gohel, P., Singh, P., Mohanty, M.: Explainable ai: current status and future directions (2021), \url{https://arxiv.org/abs/2107.07045}

\bibitem{han2015learning}
Han, S., Pool, J., Tran, J., Dally, W.J.: Learning both weights and connections for efficient neural networks. In: Advances in Neural Information Processing Systems (NeurIPS) (2015)

\bibitem{hassibi1993second}
Hassibi, B., Stork, D.G.: Second order derivatives for network pruning: Optimal brain surgeon. In: Advances in Neural Information Processing Systems (NeurIPS). pp. 164--171 (1993)

\bibitem{hatefi2024pruningexplainingrevisitedoptimizing}
Hatefi, S.M.V., Dreyer, M., Achtibat, R., Wiegand, T., Samek, W., Lapuschkin, S.: Pruning by explaining revisited: Optimizing attribution methods to prune cnns and transformers (2024), \url{https://arxiv.org/abs/2408.12568}

\bibitem{he2016deep}
He, K., Zhang, X., Ren, S., Sun, J.: Deep residual learning for image recognition. Proceedings of the IEEE conference on computer vision and pattern recognition pp. 770--778 (2016)

\bibitem{holzinger2017we}
Holzinger, A., Biemann, C., Pattichis, C.S., Kell, D.B.: What do we need to build explainable ai systems for the medical domain? arXiv preprint arXiv:1712.09923  (2017)

\bibitem{Kumar2022-gg}
Kumar, S.: {Covid19-Pneumonia-normal} chest {X-Ray} images (2022)

\bibitem{litjens2017survey}
Litjens, G., Kooi, T., Bejnordi, B.E., Setio, A.A., Ciompi, F., Ghafoorian, M., van~der Laak, J.A., van Ginneken, B., Sánchez, C.I.: A survey on deep learning in medical image analysis. Medical image analysis  \textbf{42},  60--88 (2017)

\bibitem{liu2019rethinking}
Liu, Z., Sun, M., Zhou, T., Huang, G., Darrell, T.: Rethinking the value of network pruning. International Conference on Learning Representations (ICLR)  (2019)

\bibitem{Montavon2019LayerWiseRP}
Montavon, G., Binder, A., Lapuschkin, S., Samek, W., M{\"u}ller, K.: Layer-wise relevance propagation: An overview. In: Explainable AI (2019), \url{https://api.semanticscholar.org/CorpusID:202579539}

\bibitem{montavon2017explaining}
Montavon, G., Lapuschkin, S., Binder, A., Samek, W., Müller, K.R.: Explaining nonlinear classification decisions with deep taylor decomposition. Pattern Recognition  \textbf{65},  211--222 (2017)

\bibitem{pogorelov2017kvasir}
Pogorelov, K., Eskeland, M., Tveit, K.S., de~Lange, T., Johansen, D., Espeland, H., Heslinga, C.G., Gori, C., Eskeland, S.L., Riegler, M., et~al.: Kvasir: A multi-class image dataset for computer aided gastrointestinal disease detection. In: Proceedings of the 8th ACM on Multimedia Systems Conference. pp. 164--169 (2017)

\bibitem{rajpurkar2017mura}
Rajpurkar, P., Irvin, J., Zhu, K., Yang, B., Mehta, H., Duan, T., Ding, D., Bagul, A., Langlotz, C., Shpanskaya, K., Lungren, M., Ng, A.: Mura: Large dataset for abnormality detection in musculoskeletal radiographs. arXiv preprint arXiv:1712.06957  (2017)

\bibitem{sabih2020utilizing}
Sabih, M., Hannig, F., Teich, J.: Utilizing explainable ai for quantization and pruning of deep neural networks. ArXiv  \textbf{abs/2008.09072} (2020), \url{https://api.semanticscholar.org/CorpusID:221186873}

\bibitem{sanh2020movement}
Sanh, V., Wolf, T., Rush, A.M.: Movement pruning: Adaptive sparsity by fine-tuning. In: Advances in Neural Information Processing Systems. vol.~33, pp. 20378--20389 (2020)

\bibitem{sankarapu2025dlbacktracemodelagnosticexplainability}
Sankarapu, V.K., Chitroda, C., Rathore, Y., Singh, N.K., Seth, P.: Dlbacktrace: A model agnostic explainability for any deep learning models (2025), \url{https://arxiv.org/abs/2411.12643}

\bibitem{shen2017deep}
Shen, D., Wu, G., Suk, H.I.: Deep learning in medical image analysis. Annual Review of Biomedical Engineering  \textbf{19},  221--248 (2017)

\bibitem{simonyan2014deep}
Simonyan, K., Zisserman, A.: Very deep convolutional networks for large-scale image recognition. arXiv preprint arXiv:1409.1556  (2014)

\bibitem{sundararajan2017axiomatic}
Sundararajan, M., Taly, A., Yan, Q.: Axiomatic attribution for deep networks. In: Precup, D., Teh, Y.W. (eds.) Proceedings of the 34th International Conference on Machine Learning, {ICML} 2017, Sydney, NSW, Australia, 6-11 August 2017. Proceedings of Machine Learning Research, vol.~70, pp. 3319--3328. {PMLR} (2017)

\bibitem{Sundararajan2017AxiomaticAF}
Sundararajan, M., Taly, A., Yan, Q.: Axiomatic attribution for deep networks. In: International Conference on Machine Learning (2017), \url{https://api.semanticscholar.org/CorpusID:16747630}

\bibitem{zhu2018prune}
Zhu, M., Gupta, S.: To prune, or not to prune: exploring the efficacy of pruning for model compression. In: International Conference on Learning Representations (ICLR) (2018)

\end{thebibliography}
\newpage
\appendix
\section{Baseline Model Training}
To ensure a consistent and fair evaluation across different pruning strategies and attribution methods, we first train baseline models on each medical imaging dataset without any pruning. These models serve as performance anchors for evaluating the effectiveness of our interpretability-aware pruning framework.

We applied standard augmentations during training to improve generalization. These include random resized cropping with a scale between 0.7 and 1.0 and an aspect ratio between 0.75 and 1.33, random horizontal and vertical flipping, random rotation up to 30 degrees, and color jitter with moderate brightness, contrast, and saturation variations. All images are normalized using ImageNet statistics. For validation and testing phases, only resizing, and normalization are applied to ensure consistent evaluation. To further refine the training dynamics, a step learning rate scheduler is used, which decays the learning rate by a factor of 0.1 every 10 epochs.

The training of all baseline models is carried out with fixed cross-entropy loss function with class weights to address class imbalance in the datasets. For optimization, we use the Adam optimizer, which is well-suited for training deep neural networks due to its adaptive learning rate mechanism. We train three model architectures: VGG19, ResNet50, and Vision Transformer (ViT-B/16) on each dataset using a batch size of 32 and learning rate of 0.001.

The hyperparameters used for these models represent standard performant configurations commonly used in deep learning literature and practice. They are not the result of exhaustive hyperparameter tuning.

\section{Sampling Techniques}
\label{sec:baseline}
\begin{figure}[pt]
    \centering
    \includegraphics[width=0.9\linewidth, height=\textheight, keepaspectratio]{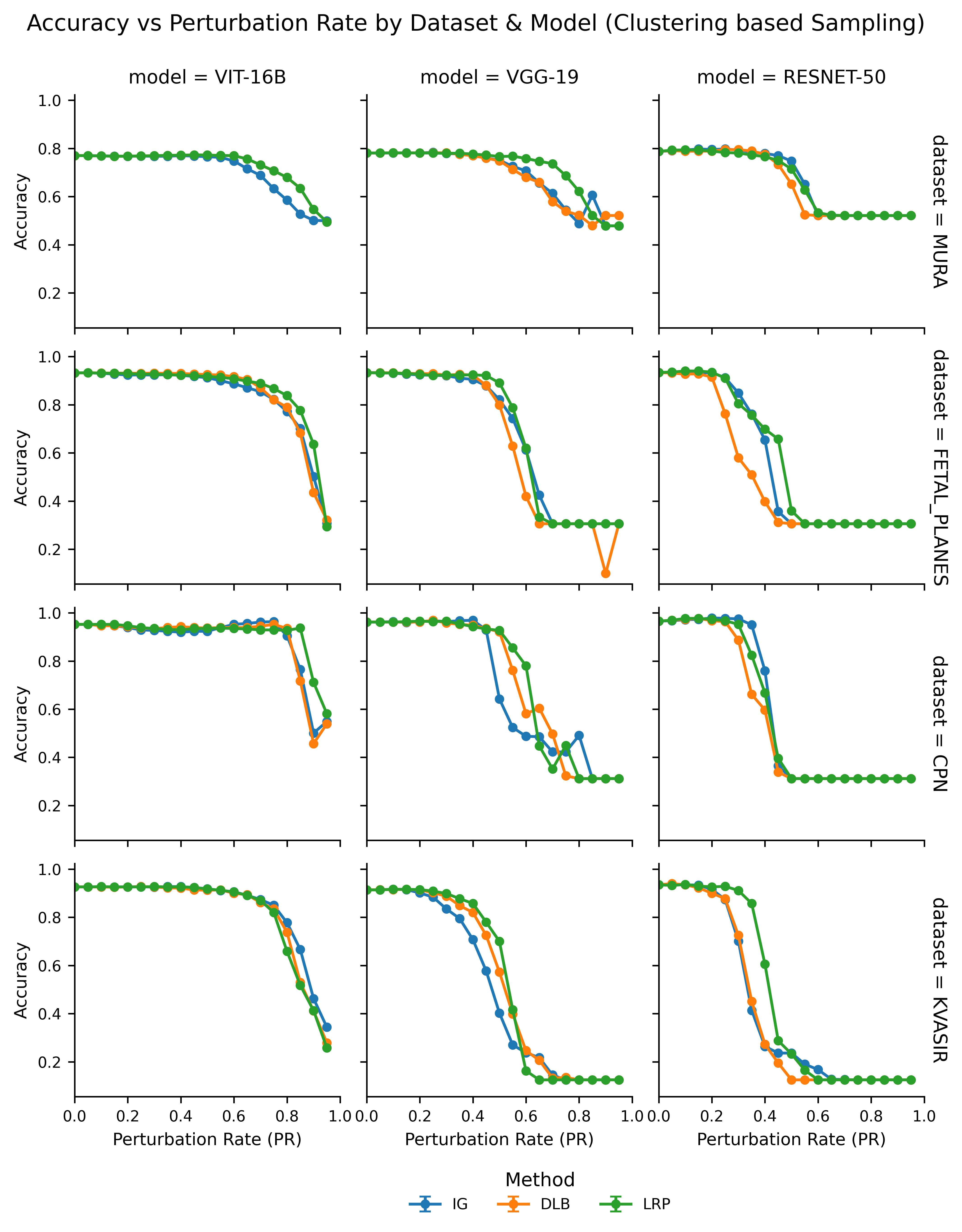}
    \caption{Accuracy vs Pruning Rate across various Models (Vit,Resnet,Vgg) and Datasets (Mura,Fetal Planes,Cpn,Kvasir) using Various Attribution Methods (IG,DLB,LRP) for clustering sampling.}
    \label{fig:pruning_clustering}
\end{figure}

\begin{figure}[pt]
    \centering
    \includegraphics[width=0.9\linewidth, height=\textheight, keepaspectratio]{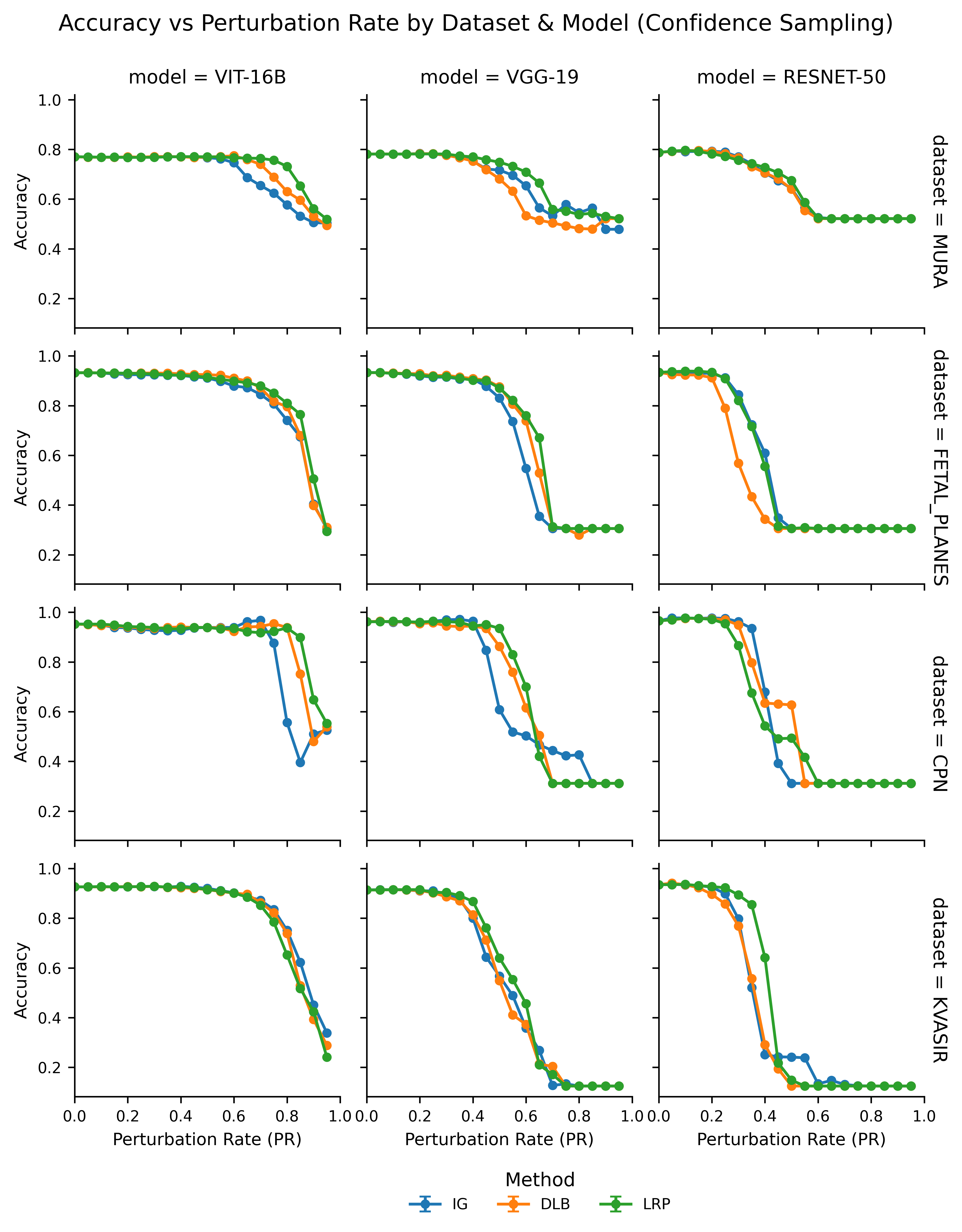}
    \caption{Accuracy vs Pruning Rate across various Models (Vit,Resnet,Vgg) and Datasets (Mura,Fetal Planes,Cpn,Kvasir) using Various Attribution Methods (IG,DLB,LRP) for confidence sampling.}
    \label{fig:pruning_confidence}
\end{figure}

Figures~\ref{fig:pruning_clustering} and \ref{fig:pruning_confidence} visualize the pruning robustness of three deep learning architectures — ViT-16B, VGG-19, and ResNet-50,  across four medical imaging datasets: Mura, Fetal Planes, Cpn, and Kvasir. The pruning is guided by three attribution methods: Integrated Gradients (IG), DL Backtrace (DLB), and Layer-wise Relevance Propagation (LRP). For each setup, the model performance is plotted against increasing pruning (perturbation) rates using different sample selection strategies for relevance computation.

\subsection{Clustering based Sample Selection}
Our experiments demonstrate that clustering-based sampling provides a robust and diverse selection of samples for computing relevance scores, outperforming random sampling in several configurations. Notably, clustering shows consistent improvements on the CPN and MURA datasets, particularly when paired with CNN-based architectures (VGG and ResNet). For instance, in the CPN-VGG setting, the use of DLB and LRP with clustering results in a higher allowable pruning threshold while preserving accuracy (e.g., 55\% vs. 45–50\% for random sampling). Similarly, in MURA with ResNet and LRP, clustering improves the 5\% accuracy drop pruning rate from 45\% to 50\%.

These improvements highlight that clustering enables the selection of semantically diverse samples, which likely represent broader decision boundaries of the model. This diversity appears to facilitate a more generalized relevance map, better identifying globally redundant neurons. The aggregated relevance obtained from diverse clusters thus offers better pruning signals, especially for architectures where activations are spatially.

However, clustering is not universally superior. Its gains are less prominent in ViTs. One of the key observations is that random sampling alone yields strong results in ViTs, with high tolerance to pruning even up to 80–85\% in some cases. The lack of significant gains from confidence or clustering-based sampling in ViTs should therefore not be interpreted as ineffectiveness of these methods, but rather as an indication that random sampling is already sufficient in transformer architectures due to their inherent representational richness and redundancy.

\subsection{Confidence based Sample Selection}

Confidence-based sampling, which selects samples with the highest softmax confidence, offers moderate gains over random sampling, particularly for CPN and Kvasir datasets. This strategy leads to slightly higher pruning tolerance in these cases, for example, in CPN-ResNet using IG (40\% vs. 30\%) and CPN-VGG with LRP (55\% vs. 50\%). The rationale behind these improvements likely stems from the fact that high-confidence samples reflect clear decision boundaries, making it easier to identify which neurons contribute most directly to decisive predictions.

However, this approach comes with inherent limitations. High-confidence predictions often correspond to "easy" samples, which may not sufficiently activate edge-case or ambiguous decision regions. Consequently, the relevance scores derived from these samples may not capture neurons that are critical in low-confidence, high-uncertainty regions—those that may be disproportionately affected during aggressive pruning. This aligns with our observation that confidence-based sampling does not consistently outperform random sampling across all datasets, and in several cases (especially with ViTs and fetal data), performs similarly or worse.

While confidence-based sampling provides an efficient and task-aligned way to guide relevance estimation, its benefits are more modest and less generalizable compared to clustering. It performs well in scenarios with clearly separable classes or well-calibrated classifiers (e.g., CPN), but may miss neurons activated by harder examples, limiting its effectiveness in more complex or noisy datasets.

\end{document}